\documentclass[10pt,twocolumn,letterpaper]{article}

\usepackage{cvpr}              

\usepackage{newtxtext}
\usepackage{amssymb, bm, bbm, mathtools,physics}
\usepackage[scr=boondoxo]{mathalfa}
\usepackage[usenames,dvipsnames,svgnames,table]{xcolor}
\usepackage[font=small]{caption}
\usepackage{float, colortbl, tabularx, longtable, multirow, subcaption, environ, wrapfig, textcomp, nicefrac}
\usepackage{booktabs,colortbl}
\usepackage{xargs}
\usepackage[T1]{fontenc}
\usepackage[final]{microtype}

\definecolor{cvprblue}{rgb}{0.21,0.49,0.74}
\usepackage[pagebackref,breaklinks,colorlinks,citecolor=cvprblue]{hyperref}
\usepackage[capitalize,nameinlink,noabbrev]{cleveref}

\definecolor{myred}{RGB}{153,51,51}
\definecolor{mygray}{RGB}{102,102,102}
\definecolor{darkgreen}{RGB}{60,160,60}
\definecolor{lightblue}{RGB}{60,60,200}
\definecolor{darkred}{RGB}{70,10,10}
\definecolor{darkblue}{RGB}{1,1,255}
\definecolor{yellow}{RGB}{255,255,0}
\definecolor{purple}{RGB}{127,0,255}
\definecolor{dark}{RGB}{1,1,1}
\definecolor{lightgray}{RGB}{230,230,230}
\definecolor{ucla_gold}{RGB}{255,232,0}
\definecolor{ucla_blue}{RGB}{50,132,191}
\definecolor{lightyellow}{rgb}{1.0, 1.0, 0.88}

\colorlet{mylinkcolor}{violet}
\colorlet{mycitecolor}{YellowOrange}
\colorlet{myurlcolor}{Aquamarine}

\definecolor{color_skyblue}{rgb}{0.01,0.39,0.75}

\usepackage[color=gray!30,textsize=tiny]{todonotes}

\def\paperID{*****} 
\def\confName{CVPR}
\def\confYear{2024}

\usepackage{caption}
\usepackage{subcaption}

\usepackage[para]{threeparttable}
\usepackage{tablefootnote}
\newcommand\blfootnote[1]{%
  \begingroup
  \renewcommand\thefootnote{}\footnote{#1}%
  \addtocounter{footnote}{-1}%
  \endgroup
}

\Crefformat{figure}{#2Fig.~#1#3}
\Crefmultiformat{figure}{Figs.~#2#1#3}{ and~#2#1#3}{, #2#1#3}{ and~#2#1#3}

\title{From NeRFs to Gaussian Splats, and Back}
\author{Siming He*, Zach Osman*, Pratik Chaudhari
}

\begin{document}


\twocolumn[{
\begin{center}
{\Large \textbf{From NeRFs to Gaussian Splats, and Back}}\\[0.75em]
Siming He*, Zach Osman*, Pratik Chaudhari\\[2em]
\end{center}
}]

\begin{figure}
\centering
\includegraphics[width=\linewidth]{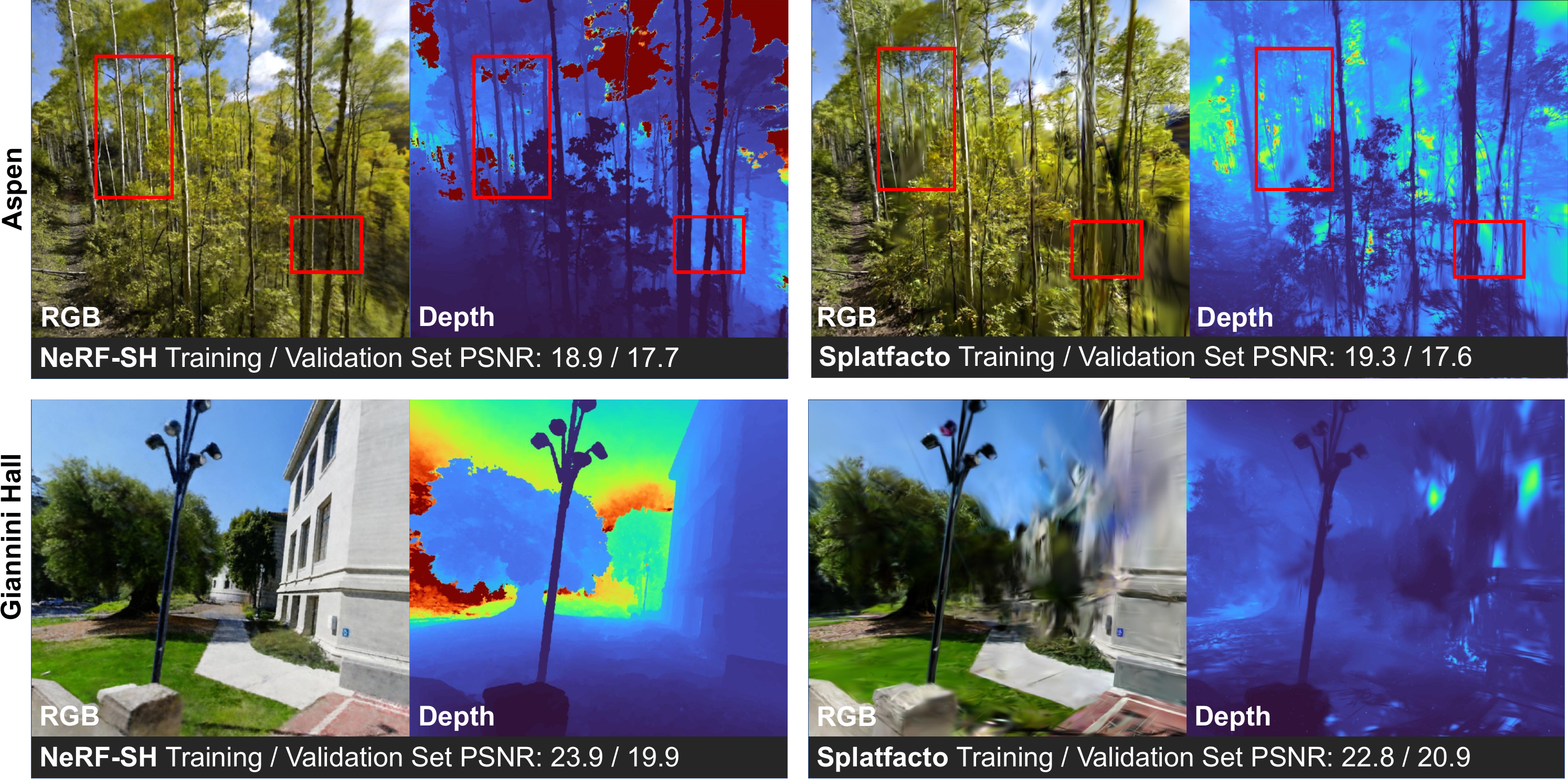}
\caption{
\textbf{NeRF generalizes better than Gaussian splatting (GS) to views that are very different from those in the training data.} 
NeRF-SH (Left) and Splatfacto (Right), both summarized in \cref{tab:models}, are trained on Aspen (Top) and Giannini Hall (Bottom) from Nerfstudio dataset~\cite{nerfstudio}.  In both datasets, NeRF-SH and Splatfacto have relatively good training and validation PSNRs to one another because the validation views are similar to the training views (see~\cref{fig:newdataset} Top). For novel views which differ more from the training views, like the images shown above, NeRF-SH renders better RGB and depth images than Splatfacto. The red boxes in the Aspen novel view illustrate areas in the RGB and depth views where NeRF-SH has noticeably better depth geometry and fewer artifacts than Splatfacto. For the Giannini Hall novel view, NeRF-SH clearly preserves the depth structure  better than Splatfacto.
}
\label{fig:result_new_1}
\includegraphics[width=\linewidth]{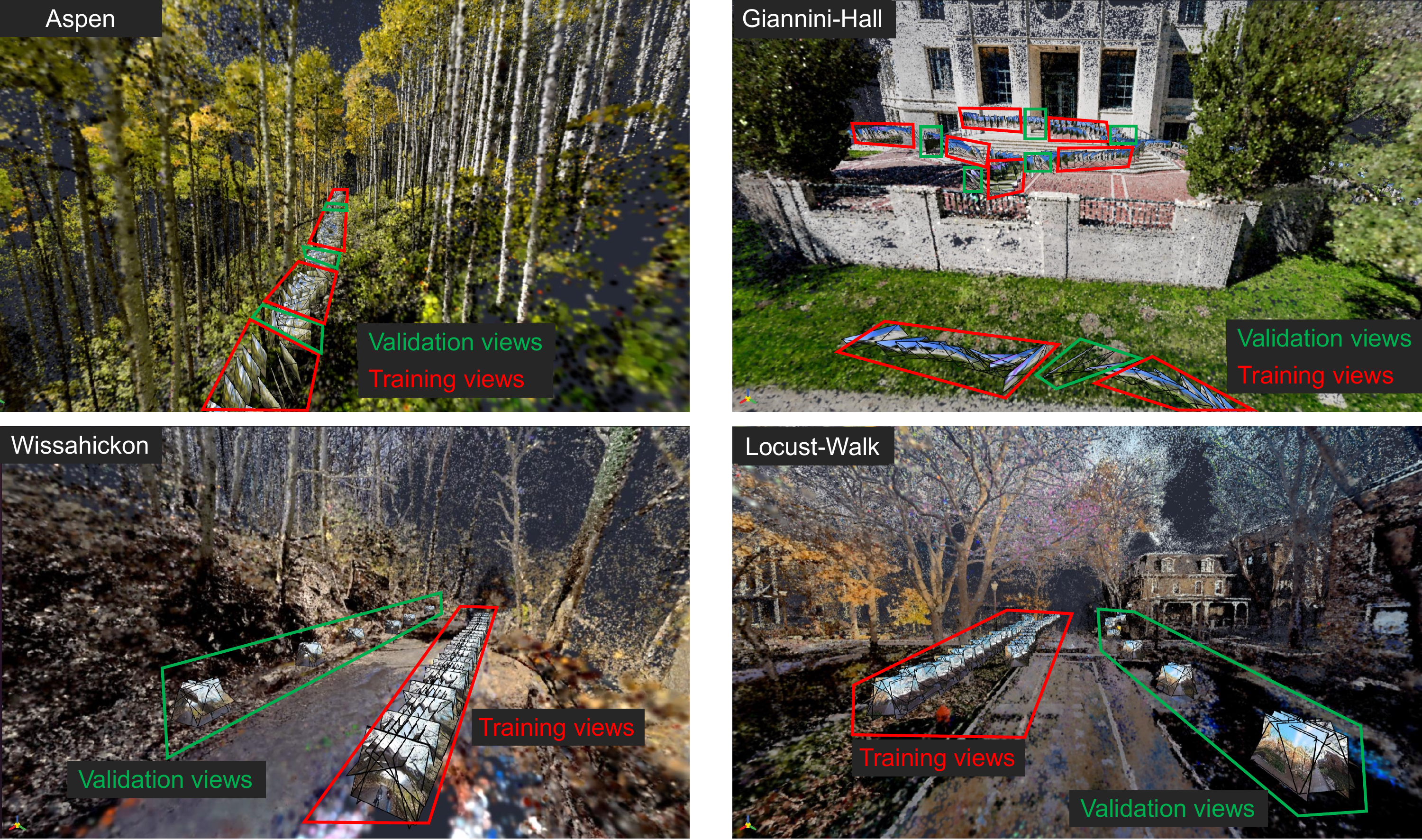}
\caption{Top: Aspen and Giannini Hall scenes from Nerfstudio~\cite{nerfstudio} contain validation views that are similar to training views. Although Splatfacto has good rendering quality on these data in~\cref{tab:results}, it does not accurately render dissimilar views in~\cref{fig:result_new_1}. Bottom: We created two new scenes, named Wissahickon and Locust Walk, where validation views are much more different than training views. For these new scenes, Splatfacto has a much worse validation PSNR than NeRF; see~\cref{tab:results}.
All images above are rendered using Gaussian splats obtained from NeRF-SH without fine-tuning; this is called NeRFGS with zero iterations in~\cref{tab:results}.
}
\label{fig:newdataset}
\end{figure}


Radiance\blfootnote{*Equal Contribution. General Robotics, Automation, Sensing and Perception (GRASP) Laboratory, University of Pennsylvania. Email: \{siminghe, osmanz, pratikac\}@seas.upenn.edu}\blfootnote{Code and data are in \href{https://github.com/grasp-lyrl/NeRFtoGSandBack}{https://github.com/grasp-lyrl/NeRFtoGSandBack}.}field-based scene representations are useful in robotics for localization and mapping~\cite{Sucar_2021_ICCV, 10341922, huang2024photoslam, yugay2023gaussianslam, keetha2024splatam, matsuki2024gaussian, yan2024gsslam}, planning and control~\cite{9712211, he2024active, liu2024uncertainty, lei2024gaussnav, li20213d}, scene understanding~\cite{shen2023F3RM, yang2023emernerf}, and simulation~\cite{byravan2022nerf2real, cleach2023differentiable, xu2024customizable}. Often, the key question in these applications is whether one uses an implicit representation like a neural radiance field (NeRF)~\cite{mildenhall2020nerf, mueller2022instant} or an explicit representation, like 3D Gaussian Splatting (GS)~\cite{kerbl3Dgaussians, huang20242d}. There are pros and cons for both.

\begin{figure}
\centering
        \includegraphics[width=\linewidth]{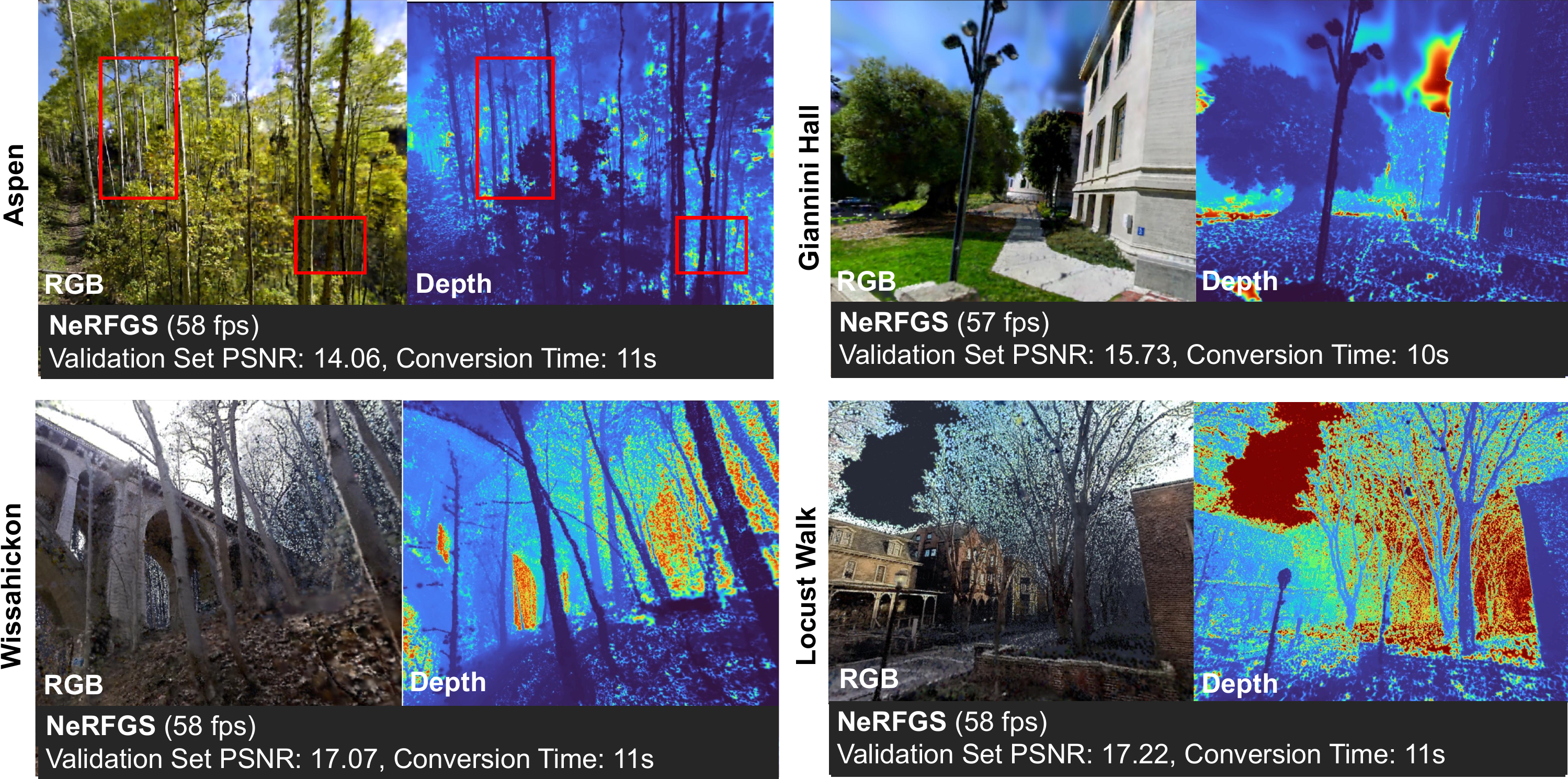}
\caption{
\textbf{NeRFGS generalizes better than GS while having real-time rendering.} NeRFGS converts trained NeRF-SH into GS while maintaining good generalization in contrast to Splatfacto in~\cref{fig:result_new_1}. The conversion takes about 10 sec on GeForce RTX 4090: 7 sec for extracting spherical harmonics and 3 sec for fine-tuning. It is therefore fast enough to be done periodically on a robot. If necessary, this time can be reduced if NERF-SH to GS conversion is done only around the robot; or if the sky (not relevant for many ground robotics tasks) is ignored.
}
\label{fig:nerf2gs}
\begin{table}[H]
\centering
\renewcommand{\arraystretch}{1.1}
\resizebox{\linewidth}{!}{
\scriptsize
    \begin{tabular}{p{1cm} p{6cm}}
        \toprule
        Nerfacto & NeRF-based approach in Nerfstudio\\
        NeRF-SH & Our modified Nerfacto that predicts spherical harmonics for the color instead of the RGB intensity\\
        Splatfacto & Gaussian Splatting approach in Nerfstudio\\
        NeRFGS & Gaussian splats obtained from NeRF-SH, with or without further fine-tuning\\
        GSNeRF & NeRF-SH converted from NeRFGS \\
        RadGS & Gaussian Splatting trained using the pointcloud obtained from NeRF~\cite{niemeyer2024radsplat} \\
        \bottomrule
    \end{tabular}
}
\caption{Different approaches for scene representations.}
\label{tab:models}
\end{table}
\includegraphics[width=\linewidth]{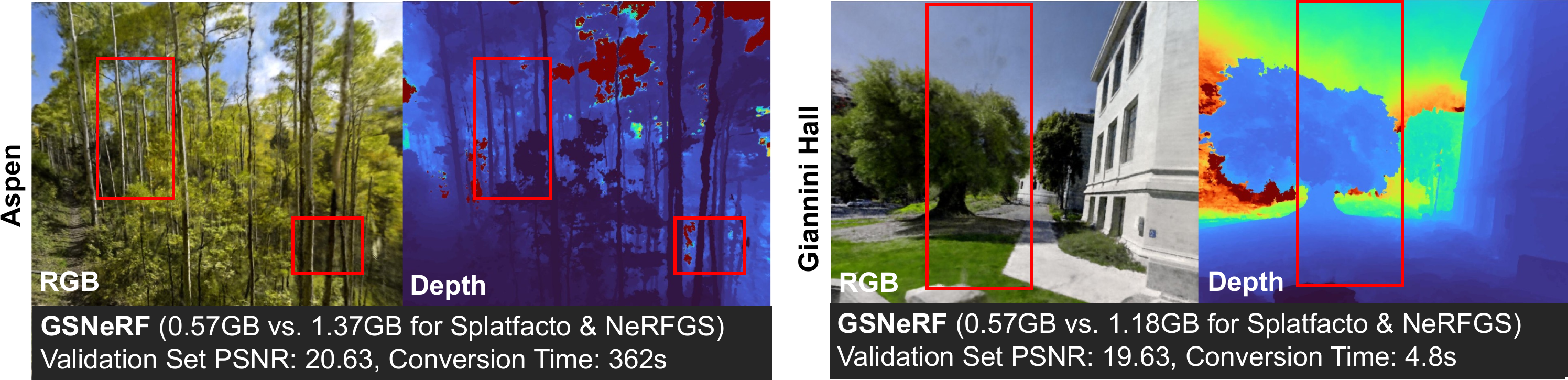}
\caption{\textbf{GSNeRF can be used for saving memory and editing the NeRF.} We convert NeRFGS model, as in~\cref{fig:nerf2gs}, back to NeRF-SH to get GSNeRF (Left). GSNeRF stores memory intensive GS models as more compact NeRFs. Also, GSNeRF with NeRFGS enables easy modifications to the scene represented by the NeRF by converting into a GS, modifying the Gaussians, and converting it back. As an example, we have edited out the lamp post from Giannini Hall in~\cref{fig:result_new_1} through GSNeRF and rendered the image (Right) using NeRF-SH.
}
\label{fig:gs2nerf}
\end{figure}

\begin{table*}
\centering
\renewcommand{\arraystretch}{1.15}
\resizebox{\linewidth}{!}{
\scriptsize
\begin{tabular}{l r | rrrrrr | rrr rrr}
\toprule
& \textbf{Iterations} & \multicolumn{3}{c}{\bf Aspen} & \multicolumn{3}{c|}{\bf Giannini Hall} & \multicolumn{3}{c}{\bf Wissahickon} & \multicolumn{3}{c}{\bf Locust Walk}  \\[0.05in]
& ($\times 10^2$) & PSNR (Val)  & SSIM & LPIPS & PSNR (Val) & SSIM & LPIPS & PSNR (Train/Val) & SSIM & LPIPS & PSNR (Train/Val) & SSIM & LPIPS \\
\midrule
Nerfacto-big~\cite{nerfstudio}  & 300     & 17.75 & 0.5  & 0.43  &  20.11     &  0.68    &   0.3    &        22.17 / 20.75          &    0.75  &    0.26   &        22.29 / 21.49          &   0.8   &   0.3    \\
Splatfacto~\cite{nerfstudio} & 300       & 17.63 & 0.5  & 0.39  & 20.87 & 0.7  & 0.33  & 23.46 / 14.62      & 0.55 & 0.45  & 24.04 / 17.72    & 0.7  & 0.31  \\
NeRF-SH & 300        & 17.73 & 0.48 & 0.45  & 19.89 & 0.65 & 0.32  & 22.41 / 17.46      & 0.61 & 0.39  & 21.73 / 18.74    & 0.7  & 0.33  \\
\midrule
RadGS~\cite{niemeyer2024radsplat} & 1 &  11.65  &  0.28  &  0.74  &  12.37    & 0.49    & 0.61  & 12.4 / 15.17  &  0.62  &  0.46  &  10.84 / 11.85 &  0.6  &  0.46 \\

RadGS~\cite{niemeyer2024radsplat} & 10 &  17.85  &  0.51  &  0.44  &  20.84    & 0.72    & 0.3  &  20.7 / 20.73   &  0.76  &  0.29  & 21.15 / 21.04 &  0.8 &  0.25  \\

\midrule
NeRFGS & 0          & 13.96  & 0.3  & 0.58  & 16.19 & 0.47 & 0.49  & - / 14.40    & 0.47 & 0.51  & - / 14.87    & 0.51  & 0.47  \\
NeRFGS & 1         & 14.06  & 0.34  & 0.57 & 15.73 & 0.53 & 0.46  & 16.62 / 17.07 & 0.63 & 0.4   &  15.7  / 17.22   & 0.65  & 0.37   \\
NeRFGS & 10           & 17.7   & 0.51  & 0.4  & 21.05 & 0.73 & 0.26  & 20.67 / 20.64 & 0.75 & 0.27  &  21.11 / 21.14   & 0.8   & 0.24 \\

\midrule
GSNeRF & 50     & 18.1 & 0.44 & 0.44  & 21.22 & 0.69 & 0.31  &   - / 17.65  & 0.63 & 0.39  &  - / 19.32   &  0.71 & 0.33 \\
GSNeRF & 300     & 18.58 & 0.51 & 0.36  & 23.71 & 0.82 & 0.17  &   - / 17.59  & 0.64  & 0.37  &  - / 19.32   & 0.72  & 0.31\\
\bottomrule
\end{tabular}
}
\caption{\textbf{Quantitative Results.}
We calculate PSNR, SSIM, LPIPS for different models and scenes. For Wissahickon and Locust Walk, where validation views are dissimilar to training views, we additionally show the training PSNR since we notice a larger gap between the training and validation PSNR for Splatfacto compared to other methods. In contrast, the rendering quality at validation views is already reasonable for NeRFGS after 100 iterations of fine-tuning. After training for 1000 iterations, NeRFGS has better quality than Splatfacto and NeRF-SH. Compared to RadGS, the convergence is faster, i.e., better results after 100 iterations, also shown in~\cref{fig:plots}. GSNeRF also trains much faster and results in better quality, see~\cref{fig:plots}.
All GS based methods can render at more than 40 FPS on GeForce RTX 4090. Training for 100 iterations takes about 6 and 3 seconds, respectively, for NeRF-based and GS-based methods. GSNeRF was supervised using the Gaussians obtained from NeRFGS with 5000 iterations.
}
\label{tab:results}
\end{table*}

Imagine a quadruped robot walking along a road. It is important to ensure that the scene representation built from its ego-centric views generalizes to new views from the other side of the road. Non-parametric GS models perform well when train are plenty and test views are similar to train views~\cite{mildenhall2020nerf, barron2022mipnerf360}. Parametric models like NeRFs work much better when train and test views are different from each other. See~\cref{fig:result_new_1}. Heuristics to move, merge, and split the Gaussians in GS are brittle for in-the-wild data with exposure variations and motion blur~\cite{niemeyer2024radsplat}. Training of NeRFs is more stable and recovers better geometry with limited views. NeRFs are also a more compact representation and require less memory than GS. This is important for resource-constrained robots. The difference is rather obvious for distilled feature fields. NeRF-based methods~\cite{shen2023F3RM, lerf2023} can store high-dimensional features efficiently. GS-based methods~\cite{shi2023language, qin2023langsplat, zuo2024fmgs} need additional steps to compress features.

Explicit representations can achieve faster rendering than implicit ones. High-speed rendering is important in robotics for localization (which requires checking many views to ascertain visual overlap with the current observation), planning (which requires synthesizing new views along putative trajectories), etc. Explicit representations can also be modified easily, e.g., by updating the Gaussians. This is useful for robots that operate in dynamic environments. Modifying implicit representations requires expensive re-training or complex modeling~\cite{park2021nerfies, pumarola2020d, li2021neural, TiNeuVox, li2023dynibar}. 

We develop a procedure to go back and forth between implicit and explicit representations. We evaluate the quality and efficiency of this approach using a number of existing datasets. We study this approach on views recorded from an ego-centric camera along hiking trails in situations when evaluation views are dissimilar to training views. We show that our approach achieves the best of both NeRFs (superior PSNR, SSIM, and LPIPS on the dissimilar views, and a compact representation) and GS (real-time rendering and ability for easily modifying the representation). The computational cost of converting between these representations is minor compared to training from scratch.

\paragraph{Results}
\cref{tab:models} provides a brief summary of the the different approaches. We modify Nerfacto to predict spherical harmonics (degree 3, i.e., 16 coefficients) for each RGB channel. The volume rendering equation remains unchanged: we calculate the RGB color from spherical harmonics using the viewing direction before integrating it along the ray.

Given such a trained ``NeRF-SH'', we calculate a point-cloud of the scene using the median depth along 2$\times 10^6$ rays rendered from training views. We ensure that these rays have high opacity and do not correspond to the sky. Isotropic Gaussians are initialized at each of these points using the density and spherical harmonics predicted by the NeRF-SH. The scale of each Gaussian is half of the average distance between each point and its three nearest neighbors. Without any further optimization, this ``NeRFGS'' already captures geometric and photometric properties of the scene impressively well; see~\cref{fig:newdataset,tab:results}. We can fine-tune it further using training views; see~\cref{fig:nerf2gs,tab:results}.

\begin{figure}[!htpb]
\centering
\includegraphics[width=0.325\linewidth]{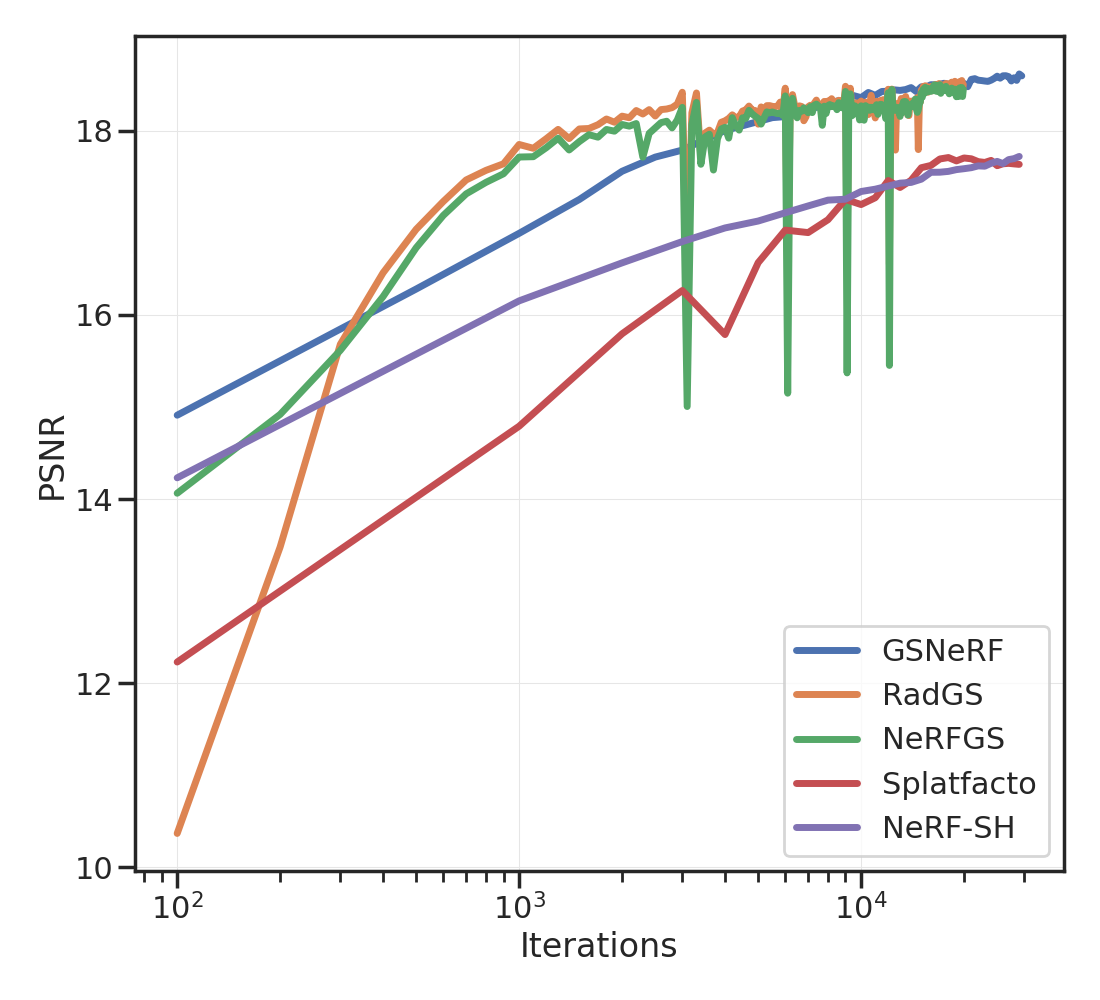}
\hfill
\includegraphics[width=0.325\linewidth]{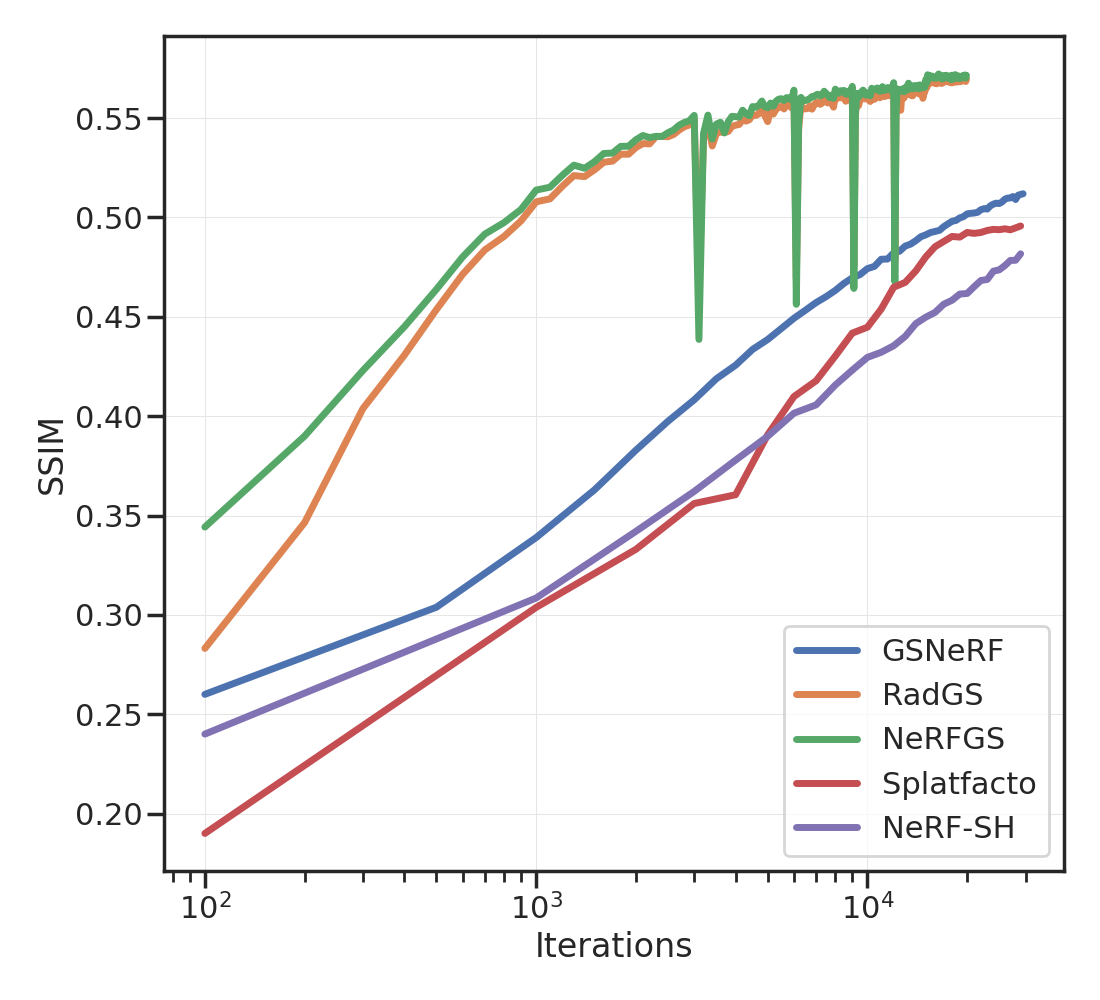}
\hfill
\includegraphics[width=0.325\linewidth]{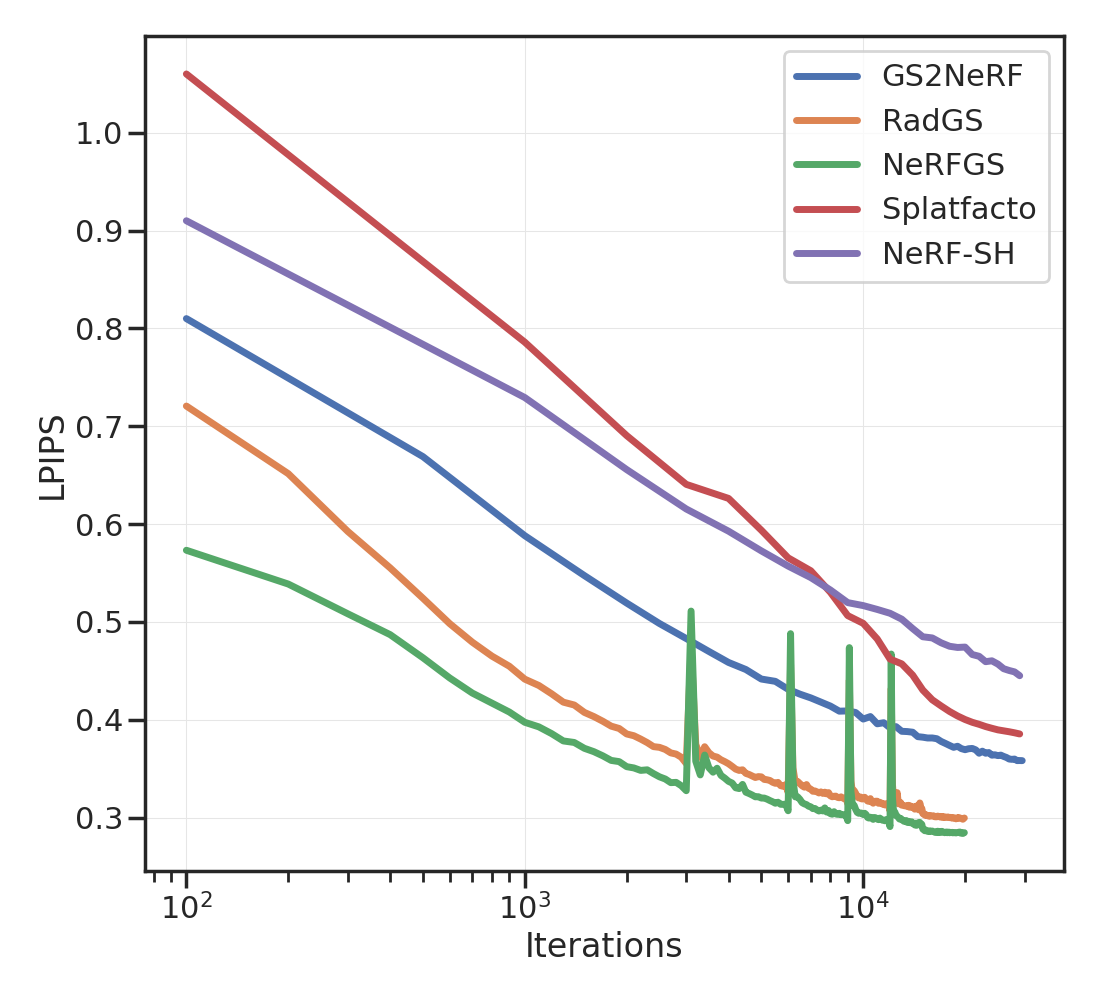}
\caption{
\textbf{NeRFs can be efficiently converted to high-quality Gaussian splats}. We report the PSNR, SSIM and LPIPS on validation data as a function of training progress for Aspen.
After 1000 iterations of fine-tuning, NeRFGS performs comparably or better than NeRF-SH and Splatfacto.
}
\label{fig:plots}
\end{figure}

For GSNeRF, we render images using NeRFGS from training views, and fit or update a NeRF-SH. We noticed that training NeRFs using GS-rendered views gives better PSNR, SSIM, and LPIPS as compared to using the original images; see~\cref{fig:plots} and~\cref{tab:results}. This is perhaps due to the absence of high-frequency structures in the GS-rendered views. One might also be interested in converting an explicit representation back into an implicit one. We show an example in~\cref{fig:gs2nerf} where we manually edit out the lamp-post by selecting the corresponding splats in NeRFGS and updating the NeRF through GSNeRF in 4.8 sec.

\paragraph{Discussion}
We demonstrated a simple procedure to convert between implicit representations of the scene such as NeRFs and explicit representations such as Gaussian splatting (GS). These ideas are useful to handle situations with sparse views, which are commonly encountered in robotics. There are many ways one might build upon this work. Notice that in~\cref{tab:results} the PSNR of NeRFGS without fine-tuning is lower than that of NeRF-SH. This indicates that there is a large degree of inefficiency in how we convert NeRF-SH into the explicit representation.

\paragraph{Acknowledgments}
This work was supported by funding from the IoT4Ag Engineering Research Center funded by NSF (NSF EEC-1941529) and NIFA grant 2022-67021-36856. 

{
    \small
    \bibliographystyle{unsrtnat}
    \bibliography{main}
}


\end{document}